\documentclass[conference]{IEEEtran}
\usepackage{cite}
\usepackage{amsmath,amssymb,amsfonts}
\usepackage{algorithmic}
\usepackage{graphicx}
\usepackage{textcomp}
\usepackage{xcolor}
\def\BibTeX{{\rm B\kern-.05em{\sc i\kern-.025em b}\kern-.08em
    T\kern-.1667em\lower.7ex\hbox{E}\kern-.125emX}}
\begin{document}

\title{Searching a Raw Video Database using Natural Language Queries\\}

\author{\IEEEauthorblockN{Sriram S K}
\IEEEauthorblockA{\textit{Dept of Computer Science Engineering} \\
\textit{PES University}\\
Bengaluru \\
sriramsk1999@gmail.com
}
\and
\IEEEauthorblockN{Siddarth Vinay}
\IEEEauthorblockA{\textit{Dept of Computer Science Engineering} \\
\textit{PES University}\\
Bengaluru \\
siddarth.vinay@gmail.com
}
\and
\IEEEauthorblockN{Srinivas K S}
\IEEEauthorblockA{\textit{Dept of Computer Science Engineering} \\
\textit{PES University}\\
Bengaluru \\
srinivasks@pes.edu
}
}

\maketitle

\begin{abstract}
The number of videos being produced and consequently stored in databases for video streaming platforms has been increasing exponentially over time. This vast database should be easily index-able to find the requisite clip or video to match the given search specification, preferably in the form of a textual query. This work aims to provide an end-to-end pipeline to search a video database with a voice query from the end user. The pipeline makes use of Recurrent Neural Networks in combination with Convolutional Neural Networks to generate captions of the video clips present in the database.
\end{abstract}

\begin{IEEEkeywords}
videos, database, query, caption, pipeline, voice 
\end{IEEEkeywords}

\section{Introduction}
Videos are one of the most popular mediums of sharing information prevalent today. They can be a medium of entertainment as well as knowledge. Creating videos is an easy process as it merely requires a camera with video capabilities, a capability available to most smartphones today. Video streaming platforms assist people with  sharing their created videos with a large number of people. There are many popular video streaming platforms in use today. These videos are stored in large databases. Searching these databases is reliant on the caption provided by the user who uploaded the video. Additionally, the captions only provide surface-level information about the videos and do not provide any scene specific information. Apart from these video streaming platforms, city surveillance systems also generate a large amount of video data from their wide network of video cameras all across their respective cities. Searching for a person of interest manually across all these video streams would be a tedious and painstaking process. An automated search system for the video database would be extremely helpful in both these cases.

Generating captions for videos requires the use of deep neural networks to create a video captioning model. A deep convolutional network is utilized to generate the set of features for a given frame or set of frames. The set of features over a few frames is fed to a recurrent neural network which then generates a corresponding sequence of words, i.e. a caption. This video captioning model was trained using the publicly available MS-COCO\cite{mscoco} dataset.

This tool aims to provide an end-to-end pipeline which can take a voice query from the end user, through the interface of a web application, search the video database to find clips of videos whose descriptions are similar to the given query and return those clips to the user. Notably, this requires certain preprocessing of the video database. First, there is a need to split each video into its constituent clips with the help of a scene detection algorithm. This is done because captions are  generated for each significant scene in a video, and therefore videos are split into smaller chunks. Next, captions are generated for each of the clips and stored as part of the database. These captions are accessed during the search process in order to return the videos relevant to the given search query.

\begin{figure}[h]
\centering
\includegraphics[width=0.45\textwidth]{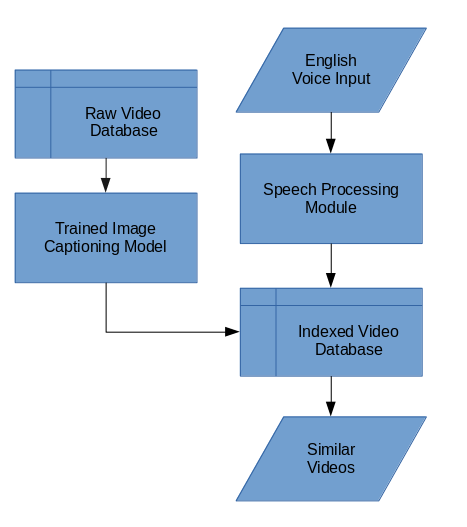}
\caption{Architecture Overview}
\label{fig:arch}
\end{figure}

The entire system is available to the end user in the form of a web application. This makes it easy and convenient to access from any device. The user can click a simple button to provide their voice input which is converted into a textual query used to search the database. The application then returns the top-3 videos matching their search query and can view the videos on the web browser. 

\section{Previous Work}

In recent years, there has been a large amount of work in the image captioning, and by extension, the video captioning fields. The goal is to generate a human readable caption which describes the objects in an image, what actions are taking place and so on. 

Li et al. \cite{persearch} describes the process by which a description of a person and their clothing is matched with an appropriate image. Wang et. al\cite{skeletonkey} takes a novel approach by generating captions in a more human fashion, identifying nouns and then attaching adjectives and verbs to those nouns to form a coherent caption. 

Vinyals et. al\cite{showtell} and Xu et. al\cite{showtellattend} however, utilize an encoder-decoder model, where the image is encoded using a CNN, and this encoding is fed into an LSTM to generate a caption. The latter paper enhances this process by making use of "attention", where the LSTM focuses on specific parts of an image to generate better captions.

In the area of video captioning, Venugopalan et. al\cite{vid2text} describes a methodology for captioning short clips of video. Frames from the video are taken and encoded into vectors, these vectors are then \textit{meanpooled} to generate a single fixed-length vector which is then fed into an LSTM to generate the video caption.

\section{Video Captioning}

The first step in building the application is the image captioning model. Since the video captioning model is simply the image captioning model with preprocessing on the input, the architecture and training process of the image captioning model is described here.

\subsection{Dataset}

\begin{figure}[h]
\centering
\includegraphics[width=0.45\textwidth]{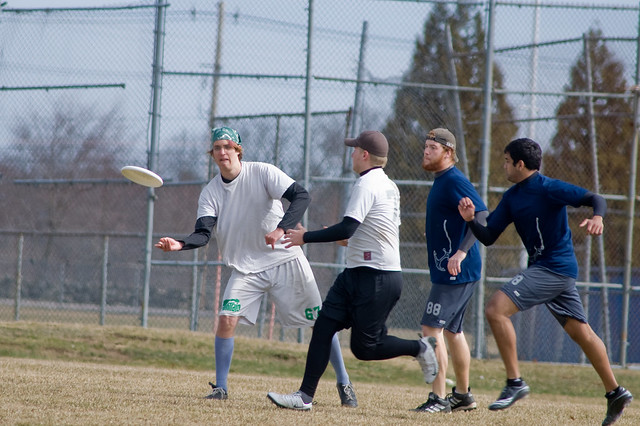}
\caption{Caption: \textit{four men are playing frisbee near a fenced-off area}}
\label{fig:sample1}
\end{figure}

\begin{figure}[h]
\centering
\includegraphics[width=0.45\textwidth]{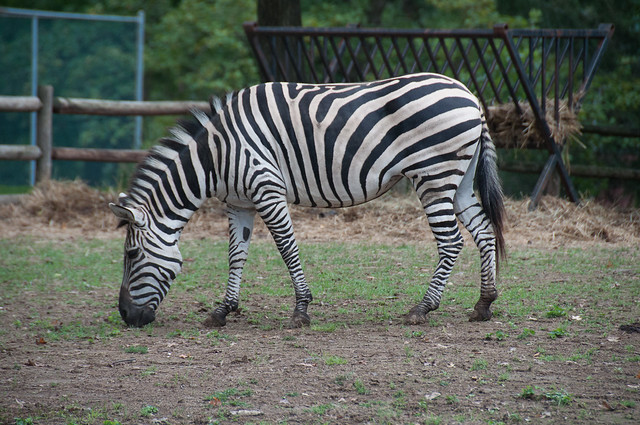}
\caption{Caption: \textit{a zebra grazing in a dirt and grass covered enclosure}}
\label{fig:sample2}
\end{figure}

The dataset being used to train the model is MS-COCO, Microsoft's Common Objects in COntext dataset, a large dataset of images with each image being heavily annotated with additional information such as segmentation, object detection and of course, captions. Samples from the dataset are shown in Fig. \ref{fig:sample1} and Fig. \ref{fig:sample2}. The model is trained on a subset of the dataset, around 250k images, due to hardware limitations.

\subsection{Model}

The model being used is a variation of the encoder-decoder model. These models are commonly used in machine translation tasks, and image captioning is considered a machine translation task since conversion occurs from one domain (images) to another (text).

In an encoder-decoder model, the image is \textit{encoded} into a fixed-length vector, which is then fed into a decoder which \textit{decodes} the vector into a variable length output. In the case of image captioning, the encoding is done by a CNN which transforms an image into an encoded vector, which is then fed into the decoder, an LSTM which is enhanced with attention.

The encoder is a convolutional neural network, specifically it is the popular CNN InceptionV3 network, pretrained on the ImageNet\cite{imagenet} dataset. The principle of transfer learning is utilized here, where a network trained on one task can be re-purposed for a similar task. The last layer of the network (the layer which classifies objects in the original network) is removed and instead a final layer encoding layer is added.

The decoder is a Long Short Term Memory (LSTM) network, a popular variant of a recurrent neural network, modified to mitigate the problems plaguing the latter like vanishing gradients. As shown in Xu et. al, \textit{attention} is utilized to better performance. When using attention, a \textit{context vector} that ascribes weights to the input is generated. This allows the decoder to focus on the more important parts of the image as the context vector improves with training.

\begin{figure}[h]
\centering
\includegraphics[width=0.45\textwidth]{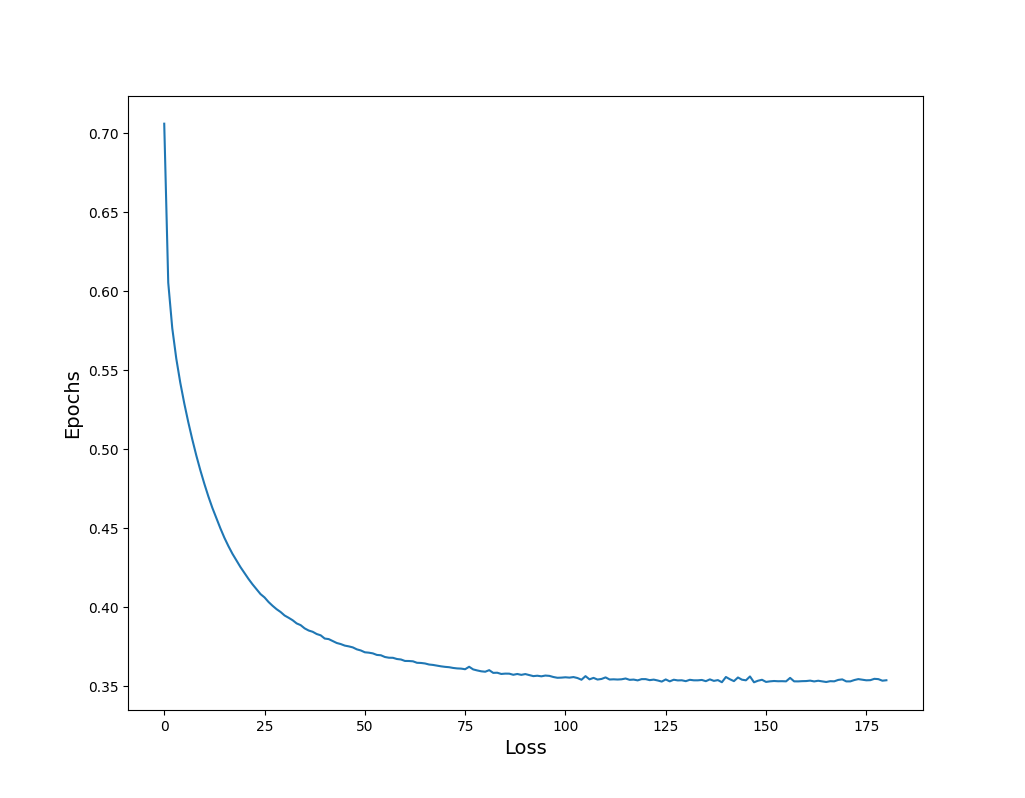}
\caption{Training loss}
\label{fig:trainloss}
\end{figure}

\begin{figure}[h]
\centering
\includegraphics[width=0.45\textwidth]{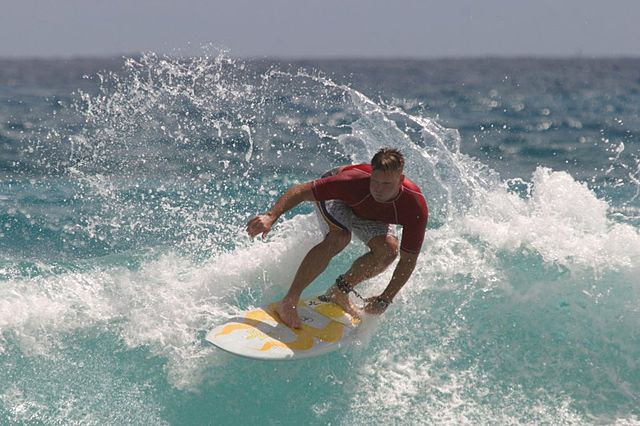}
\caption{Caption: \textit{a man holding is tray a tall on a surfboard}}
\label{fig:result1}
\end{figure}

\begin{figure}[h]
\centering
\includegraphics[width=0.45\textwidth]{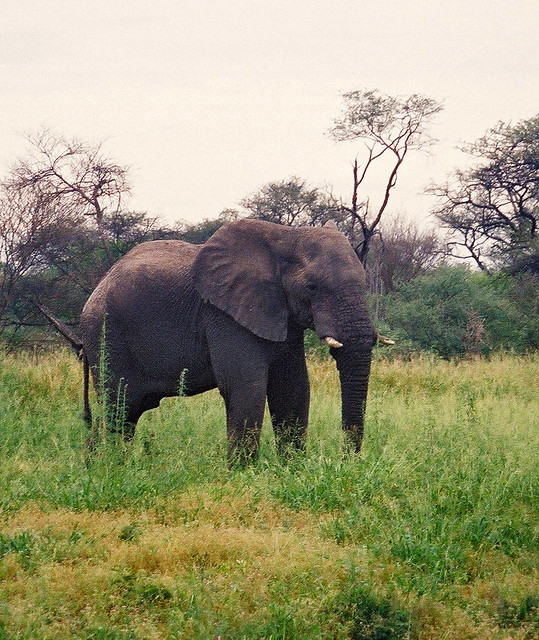}
\caption{Caption: \textit{a small elephant with a tower tennis a up of flowers huge bus tennis trees}}
\label{fig:result2}
\end{figure}

\begin{figure}[h]
\centering
\includegraphics[width=0.45\textwidth]{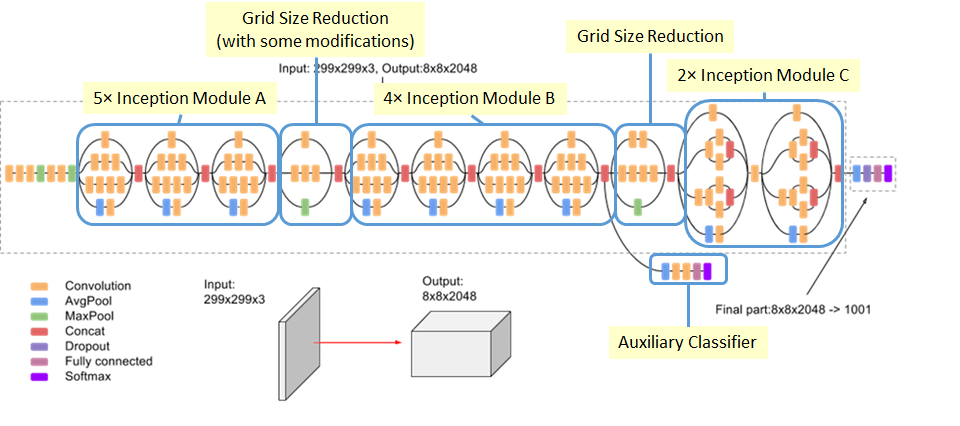}
\caption{InceptionV3 network architecture}
\label{fig:inceptionv3}
\end{figure}

\begin{figure}[h]
\centering
\includegraphics[width=0.45\textwidth]{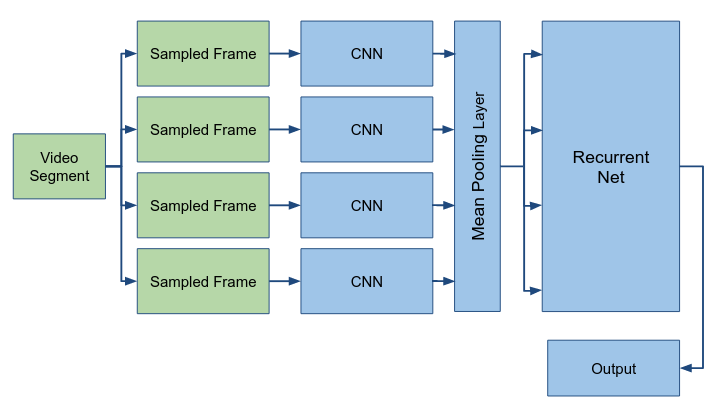}
\caption{High level look at model architecture}
\label{fig:vidcapmodel}
\end{figure}

\subsection{Training Procedure}

A brief description of the training procedure of the model follows.

\begin{itemize}
    \item Firstly, the images and the captions associated with them are loaded. Due to hardware limitations, only 250k images from the COCO dataset are used. Similarly, instead of making use of the InceptionV3 network every time, the images are run through the network once and the resultant feature vectors are stored to disk.
    \item The next step is to process the captions, by padding all small captions to the length of the maximum caption, and replacing all infrequent words with the <UNK> tag to enhance learning.
    \item Finally, both the encoder and the decoder are trained for 200 epochs. Progress is checkpointed every few epochs in case of any interruption and the the models load weights from the latest checkpoints to maintain progress.
\end{itemize}

The training loss is presented in Fig. \ref{fig:trainloss} along with some sample results in Fig. \ref{fig:result1} and Fig. \ref{fig:result2}.

\section{Scene Detection}
A scene in a video refers to a continuous sequence of frames containing an unbroken flow of action from a fixed perspective. Scene Detection refers to the process of splitting a video into its constituent scenes. It is an important preprocessing stage of the end-to-end pipeline. Scene detection is applied to each of the videos in the video database as the captioning model requires unbroken footage for captioning of clips. Additionally, splitting and captioning clips of the video instead of the entire video enables searching of more specific terms within the video. 

There are multiple methods of performing scene detection. One of these methods is to perform edge detection for each frame and comparing differences in edges\cite{scene1}. This method can handle different types of scene changes; including fades, dissolves and wipes. Another relatively new method is to perform histogram shape-based scene detection, where a 2D histogram is generated for each frame and the histogram shape is extracted\cite{scene2}. However, these methods would have required a higher processing time which would have been exacerbated by the number of videos in the database.

The method chosen was to generate the histogram for each frame within the video and comparing it with the previous frame using a common distance measure, such as the Euclidean distance between the flattened histograms of two frames. The scene detection on a sample video with standard cuts was observed taking into account multiple thresholds. It was found that a Euclidean Distance of 0.3 between the frame histograms was an adequate threshold to determine changes in scene.

The given method was implemented in Python with the help of the OpenCV library\cite{opencv}. 3D histograms of each frame was generated and the Euclidean distance was calculated for the flattened histogram using the function available as part of the SciPy Python package\cite{scipy}.

\begin{figure}[h]
\centering
\includegraphics[width=0.45\textwidth]{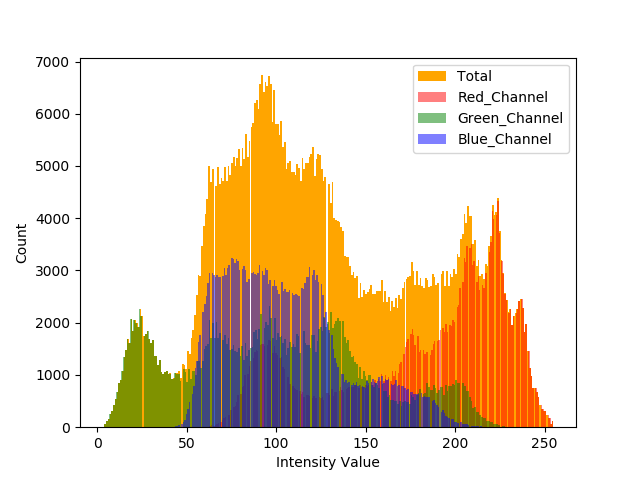}
\caption{Sample Image Histogram with Red, Green and Blue Channels }
\end{figure}

\section{Video Database}
The video database chosen to demonstrate the pipeline is the MSR-VTT (Microsoft Research - Video To Text) dataset provided by Microsoft as part of their MSR Video to Language Challenge, 2017\cite{msrvtt}. The dataset consists of a 10,000 videos which fall under any of 20 defined categories. Each of the videos also comes with 20 different natural language descriptions. As a result, this dataset is considered to be the largest dataset in terms of clip-sentence pairs (with 200,000 pairs) and in terms of vocabulary (29,316 unique words). 

The dataset does not provide the original video clips directly. The original videos are hosted on YouTube. The dataset is provided in the form of a JSON (JavaScript Object Notation) file. This file contains the videos and sentences together. The video data consists of a video ID, the URL, category and the start and end time of the video clip within the entire video. The sentence data consists of the video ID along with the caption for the video in question.

As the videos from the database are hosted on YouTube, simple preprocessing was required to verify that the videos are currently available to be accessed and downloaded. This required a simple Python script using the \textbf{youtube-dl} package to check for the availability of each of the videos. For demonstration purposes, a smaller subset of videos was chosen. First, 6 distinct categories of video were chosen: food, news, people, science, sports, travel. From each of these categories, 100 videos were downloaded for captioning. 

After the image captioning model is trained, the next step is applying the same to videos i.e. video captioning. This section presumes the existence of a number of small videos ready to be captioned. The process of video captioning is fairly straightforward. 

For each video to be captioned, the process is as follows:

\begin{itemize}
    \item The video is loaded with the help of OpenCV and every \textit{nth} frame of the video is iterated over, where $n=10$ was chosen, a reasonable number for videos whose framerates are around 30.
    \item The frame is captured, resized, and then passed through the InceptionV3 pretrained network to give us a single set of features. This is repeated for every \textit{nth} frame, finally returning a vector whose length is $l//n$ where $l$ is the total number of frames in the video and $n$ is the number of frames to skip between each iteration.
    \item The \textit{meanpool} operation is performed on the vector, collapsing it into a vector of length 1. This is analogous to collapsing a video into a single image. 
    \item Finally, this meanpooled vector is fed to image captioning model, generating a caption for the video, which is then stored in a dictionary of the form \textbf{\{filepath:caption\}}.
\end{itemize}

Finally, after all captions are generated, the dictionary is stored in a JSON file which is used for lookup after deploying the web application.

\section{Web application}

The final step is integrating all the components and deploying it as a web application. A simple web application utilizing Flask as the backend server was written. A primary goal of this tool was to create an application that was usable through voice queries alone. So, the web application takes in voice as input, using the MediaRecorder API that is built into the browser. This is sent to the backend through the use of AJAX, where it is processed using Google's voice-to-text APIs and finally, all relevant videos are returned to the user.

The web application is a RESTful application making use of REST APIs for its functionality. The user's voice query is sent to the backend, where it is processed and the query text is returned the user. Following this, the query is matched against every caption in the indexed video database and finally, the top 3 videos whose captions match the most with the query are returned to the user. More specifically, the query is compared with the caption of every video using the METEOR (Metric for Evaluation of Translation with Explicit ORdering) metric, commonly used for measuring the quality of machine translation output. In this way, the captions which have the highest meteor score with respect to the query are calculated, and the videos corresponding to that query are returned. 

\begin{figure}[h]
\centering
\includegraphics[width=0.45\textwidth]{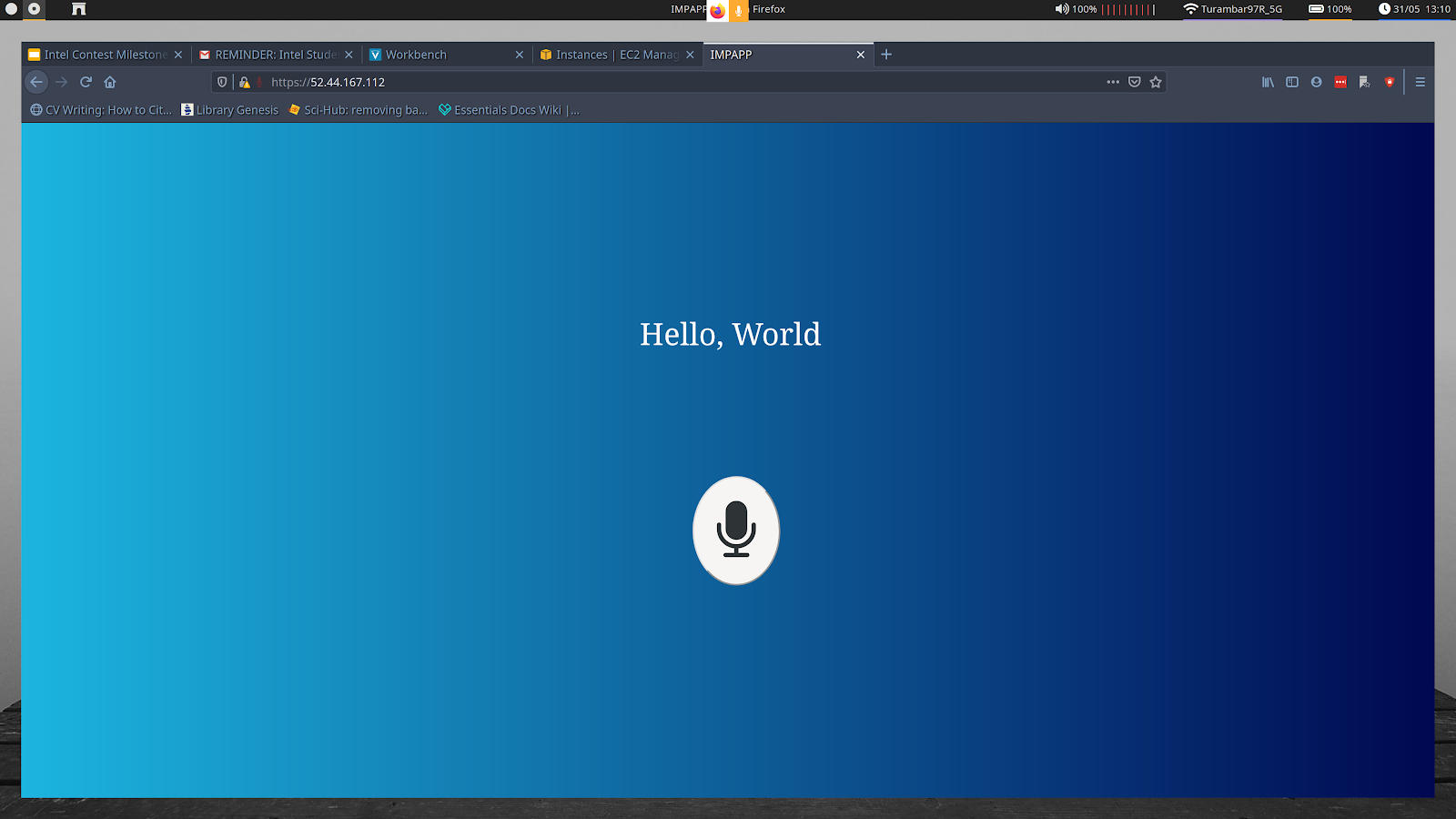}
\caption{Home Page}
\label{fig:final1}
\end{figure}

\begin{figure}[h]
\centering
\includegraphics[width=0.45\textwidth]{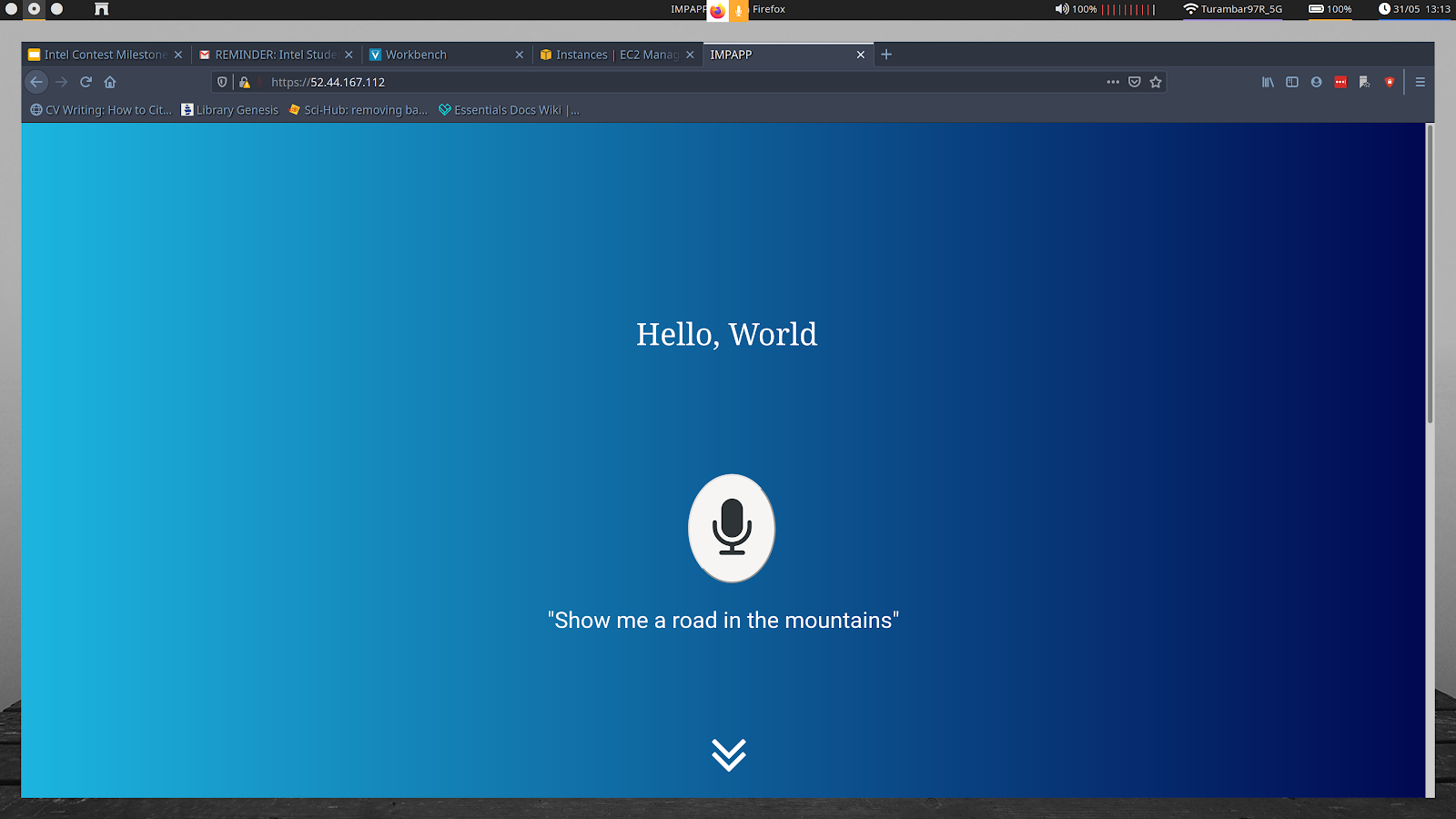}
\caption{The result of the spoken query}
\label{fig:final2}
\end{figure}

\begin{figure}[h]
\centering
\includegraphics[width=0.45\textwidth]{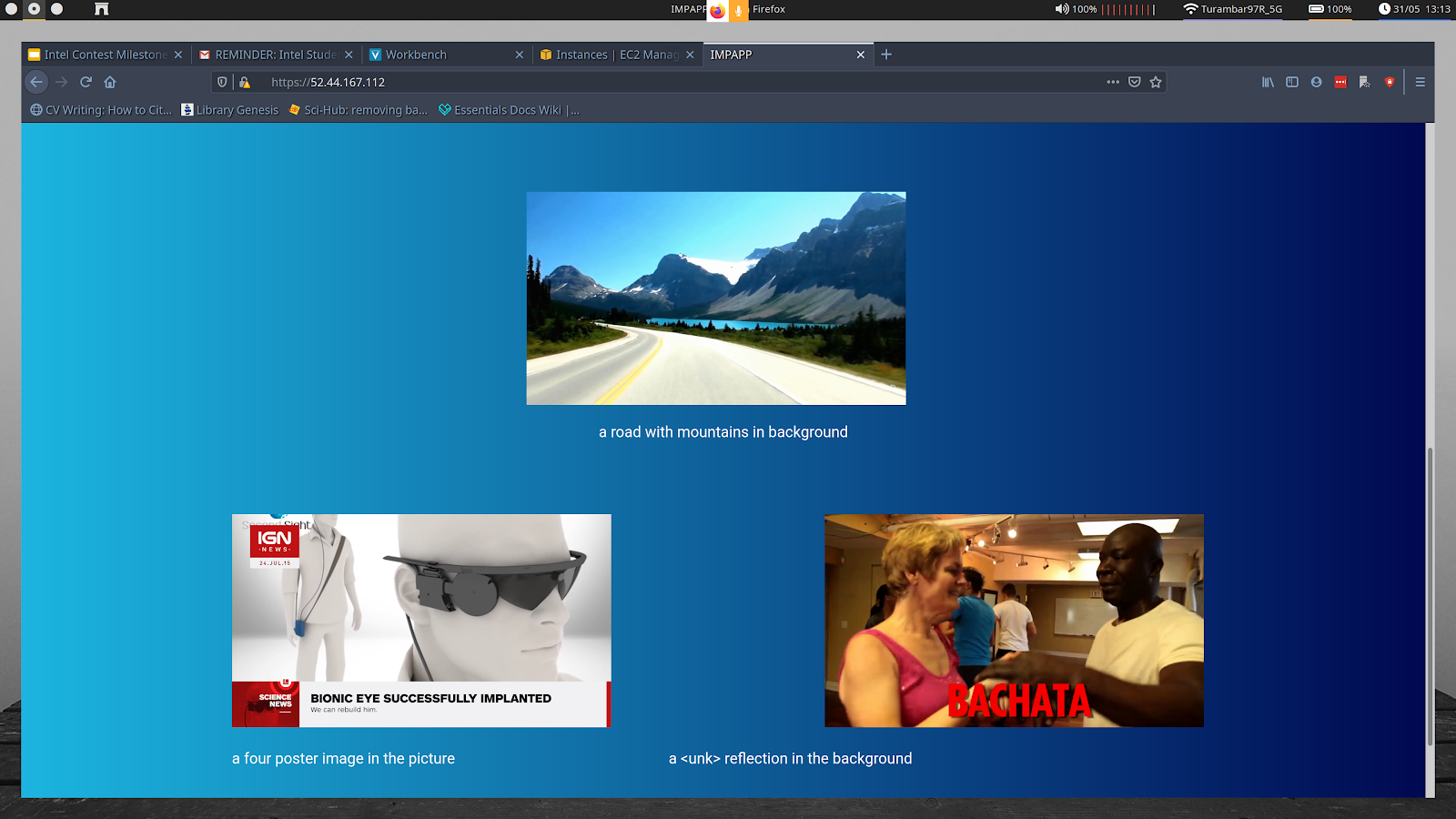}
\caption{Final output from application}
\label{fig:final3}
\end{figure}

The paths of these videos are then attached to the <video> tags in the front end using JavaScript, which triggers an asynchronous call for the videos themselves, which are fetched after a short duration. The functioning of the web app is shown Fig. \ref{fig:final1}, Fig. \ref{fig:final2} and Fig. \ref{fig:final3}. 

\section{Conclusion}

In conclusion, an end to end pipeline to search a raw video database through voice query alone is developed, and deployed as a web app for ease of use. There are numerous areas of improvement to be made and the authors hope to improve the performance of the tool over time.

The performance of the captioning model can be improved in multiple ways. Firstly, a newer approach to captioning itself. Newer research \cite{skeletonkey} suggests that computers can learn better if they compose their sentences in a more intelligent fashion, generating nouns first and then the modifiers associated with them. Other than this,  the performance of the existing model can be optimized by tuning various hyper-parameters to maximize accuracy. Finally, access to better hardware would allow the authors to train on the complete COCO dataset instead of a subset, and for more epochs in a reasonable amount of time. 

The tool can further be improved with changes to the web application. One current drawback of the current application is the requirement of a Secure HTTP (HTTPS) connection which has have implemented with the help of a self-signed SSL (Secure Sockets Layer) Certificate. This leads to most modern browsers to detect the application as unsafe. Another improvement would be to optimize the interface to be easily usable on mobile devices. Accessibility on mobile devices could also be achieved with the help of a mobile application.

\section*{Acknowledgement}

This work was done in part or full under the Intel Corporation Student Contest 2020 in association with the Center for Innovation and Entrepreneurship (CIE) at PES University. The authors acknowledge all the support from Intel Corporation towards this work.


\begin{thebibliography}{00}
\bibitem{mscoco} Lin, Tsung-Yi, et al. "Microsoft coco: Common objects in context." European conference on computer vision. Springer, Cham, 2014.
\bibitem{skeletonkey} Wang, Yufei, Zhe Lin, Xiaohui Shen, Scott Cohen, and Garrison W. Cottrell. "Skeleton key: Image captioning by skeleton-attribute decomposition." In Proceedings of the IEEE conference on computer vision and pattern recognition, pp. 7272-7281. 2017.
\bibitem{seq2seq} Venugopalan, Subhashini, Marcus Rohrbach, Jeffrey Donahue, Raymond Mooney, Trevor Darrell, and Kate Saenko. "Sequence to sequence-video to text." In Proceedings of the IEEE international conference on computer vision, pp. 4534-4542. 2015.
\bibitem{persearch} Li, Shuang, Tong Xiao, Hongsheng Li, Bolei Zhou, Dayu Yue, and Xiaogang Wang. "Person search with natural language description." In Proceedings of the IEEE Conference on Computer Vision and Pattern Recognition, pp. 1970-1979. 2017.
\bibitem{vid2text} Venugopalan, Subhashini, Huijuan Xu, Jeff Donahue, Marcus Rohrbach, Raymond Mooney, and Kate Saenko. "Translating videos to natural language using deep recurrent neural networks." arXiv preprint arXiv:1412.4729 (2014).
\bibitem{inception} Szegedy, Christian, Sergey Ioffe, Vincent Vanhoucke, and Alexander A. Alemi. "Inception-v4, inception-resnet and the impact of residual connections on learning." In Thirty-first AAAI conference on artificial intelligence. 2017.

\bibitem{showtell} Vinyals, Oriol, Alexander Toshev, Samy Bengio, and Dumitru Erhan. "Show and tell: A neural image caption generator." In Proceedings of the IEEE conference on computer vision and pattern recognition, pp. 3156-3164. 2015.
\bibitem{showtellattend} Xu, Kelvin, Jimmy Ba, Ryan Kiros, Kyunghyun Cho, Aaron Courville, Ruslan Salakhudinov, Rich Zemel, and Yoshua Bengio. "Show, attend and tell: Neural image caption generation with visual attention." In International conference on machine learning, pp. 2048-2057. 2015.

\bibitem{flask} Grinberg, Miguel. Flask web development: developing web applications with python. " O'Reilly Media, Inc.", 2018.

\bibitem{scene1} Zabih, Ramin, Justin Miller, and Kevin Mai. "A feature-based algorithm for detecting and classifying scene breaks." In Proceedings of the third ACM international conference on Multimedia, pp. 189-200. 1995.
\bibitem{scene2} Cho, Sung In, and Suk-Ju Kang. "Histogram Shape-Based Scene-Change Detection Algorithm." IEEE Access 7 (2019): 27662-27667.
\bibitem{opencv} Bradski, Gary, and Adrian Kaehler. Learning OpenCV: Computer vision with the OpenCV library. " O'Reilly Media, Inc.", 2008.
\bibitem{scipy} Virtanen, Pauli, Ralf Gommers, Travis E. Oliphant, Matt Haberland, Tyler Reddy, David Cournapeau, Evgeni Burovski et al. "SciPy 1.0: fundamental algorithms for scientific computing in Python." Nature methods 17, no. 3 (2020): 261-272.

\bibitem{msrvtt} Xu, Jun, Tao Mei, Ting Yao, and Yong Rui. "Msr-vtt: A large video description dataset for bridging video and language." In Proceedings of the IEEE conference on computer vision and pattern recognition, pp. 5288-5296. 2016.

\bibitem{imagenet} Deng, Jia, et al. "Imagenet: A large-scale hierarchical image database." 2009 IEEE conference on computer vision and pattern recognition. Ieee, 2009.

\end{thebibliography}
\end{document}